\documentclass[11pt,twocolumn]{scrartcl}

\usepackage{casa_conf}	
\usepackage{graphicx}	
\usepackage{tabularx}
\usepackage{booktabs} 
\usepackage{diagbox}  
\usepackage{multirow}  
\usepackage{caption}
\usepackage{longtable}
\usepackage{pdflscape}
\usepackage{changepage} 
\usepackage{stfloats}
\usepackage[caption=false,font=normalsize,labelfont=sf,textfont=sf]{subfig}

\title{The Detection and Rectification for Identity-Switch Based on the Unfalsified Control}

\author{Junchao Huang\\
        j584356932549@sjtu.edu.cn\\
        Shanghai Jiao Tong University\\
        \and
        Xiaoqi He*\\
        hexiaoqi@niii.com\\
        SJTU Ningbo Institute\\
        \and
        Yebo Wu\\
        yc37926@um.edu.mo\\
        Universidade de Macau\\
         \and
        Sheng Zhao\\
        shengzhao@sjtu.edu.cn\\
        Shanghai Jiao Tong University\\}

\begin{document}

\maketitle

\begin{abstract}
The goal of multi-object tracking (MOT) is to continuously track and identify objects detected in videos. Currently, most methods for multi-object tracking model the motion information and combine it with appearance information to determine and track objects. However, overlapping between different targets can still lead to identity switch issues. To meet this challenge, we propose unfctrack, which employs unfalsified control to address  the identity-switch problem in multi-object tracking. Specifically, we establish sequences of appearance information variations for the trajectories during the tracking process and a detection and rectification module is designed for identity-switch detection and recovery. Additionally, a simple and effective strategy is proposed to address the issue of ambiguous matching of appearance information during the data association process. Extensive experiments are conducted to evaluate the effectiveness of the unfctrack on the public MOT datasets. The results demonstrate that the unfctrack exhibits excellent effectiveness and robustness in handling tracking errors caused by occlusions and rapid movements. 
\end{abstract}
\linebreak
\linebreak


\section{Introduction}
Multi-object tracking \cite{ref42} (MOT) aims to track and identify the trajectories of multiple objects in a video scene. With the rapid development of object detection methods, the current mainstream and effective approaches \cite{ref17,ref7,ref3} still rely on detection-based tracking paradigms. Detection-based tracking methods transform the multi-object tracking problem into a data association problem, where the current frame detection boxes are associated with the detection boxes from the previous frame to establish object trajectories. However, errors in information acquisition, processing, and prediction during the tracking process can lead to matching errors in data association, resulting in ID-switch in multi-object tracking.Current research efforts have been focused on reducing the occurrence of ID-switch by addressing data sources and data association processes \cite{ref2,ref16},but the approaches are always to find ways to reduce the occurrence of id-switch, rather than making rectification after it occurs.

In this work, we take a different perspective on the ID-switch problem. Specifically, instead of solely reducing the occurrence of ID-switch, we also focus on identifying whether ID-switch occur and attempt to correct it. We establish a multi-object tracking model based on unfalsified control \cite{ref28}, which enables the tracker to monitor the state of objects and includes the ID-switch detection module(IDSD). The tracker incorporates historical data that is often overlooked. Additionally, we introduce the ID-switch rectification module(IDSR) based on historical information to attempt the recovery of objects that have experienced ID-switch, making our tracker the first in the MOT field to address ID-switch and attempt rectification. The Ambiguous match improvement module(AMI) in our tracker effectively reduces the problems caused by small differences in appearance information during the data association process. Moreover, the IDSD, IDSR, and AMI modules of our tracker can be easily integrated into other tracking approaches. We conduct experiments on MOT datasets and achieve promising results. We believe that our approach provides a new possibility for addressing the ID-switch problem in multi-object tracking. Finally, we discuss the limitations and applicability of our approach.

\section{Related works}
Different approaches in data processing lead to variations in multi-object tracking (MOT) methods, which can be broadly categorized into the following types: motion-based tracking, appearance-based tracking, and other types of learning-based tracking.
\subsection{Motion-based tracking}

Motion information is primarily used in the data association process. Motion-based trackers typically employ methods such as Kalman filtering  \cite{ref23} and particle filtering \cite{ref24} to predict the motion of objects and then match their predicted positions with the detection boxes in the next frame to establish data association. SORT \cite{ref17} utilizes Kalman filtering for box prediction and performs data association using IOU and the Hungarian algorithm \cite{ref25}. Bytetrack \cite{ref3} focuses on leveraging low-confidence detection boxes, while OCR-SORT \cite{ref5} emphasizes modeling motion based on observed results, improving tracking robustness in nonlinear motion scenarios. These motion models employ Bayesian estimation to predict the next state and establish tracking trajectories through data association, but they ignore the utilization of other information such as appearance information.

\subsection{Appearance-based tracking}

With the advancements in the field of re-identification(ReID), effective methods for extracting appearance features have been proposed \cite{ref40,ref21}, leading to the integration of appearance information in multi-object tracking research. DeepSORT \cite{ref7} is one of the early trackers that incorporates appearance information by combining it with motion information in the data association process, improving the robustness of tracking systems under occlusion. ATOM \cite{ref31} replaces Kalman filtering with deep learning for box estimation, while BOT-SORT \cite{ref2} incorporates ReID features into high-resolution detection boxes. Finetrack \cite{ref9} introduces a feature pyramid network that learns semantic flows between feature maps of different resolutions to correct spatial misalignments, enabling more accurate learning of appearance features. Approaches such as FairMOT \cite{ref11,ref12} perform object detection and appearance feature extraction within a single network, allowing end-to-end training and reducing inference time. More recently, Deep OCR-SORT \cite{ref6} adaptively integrates appearance matching into existing high-performance motion-based methods using object appearance, further enhancing tracking performance.The utilization of appearance information increases the robustness of the tracker in occlusion situations, but sometimes leads to fuzzy matching problems.

\subsection{Other types of learning-based tracking}

Motiontrack \cite{ref16} employs trained interaction and patrol modules to handle complex motion in dense crowds and reidentify lost trajectories. CBIOU \cite{ref4} extends the detection and tracking matching space by adding a buffer, mitigating the impact of irregular motion. MAAtrack \cite{ref10} proposes a novel tracking association method that models fuzzy matching by searching for potential track detections with similar distances. In recent years, the successful application of transformer \cite{ref26} in the visual domain, particularly the work by \cite{ref27}, has sparked a wave of combining transformer with multi-object tracking. Approaches such as \cite{ref8,ref18,ref19} treat multi-object tracking as a sequence prediction problem, where each sequence corresponds to a trajectory of an object. MOTR \cite{ref8}, in particular, achieves end-to-end multi-object tracking by tracking queries and objects. Additionally, utilizing federated learning \cite{ref43} can address data privacy issues in multi-object tracking processes.

\vspace{10pt}

In conclusion, existing methods are insufficient to rectify the occurrence of identity-switch. Therefore, we propose unfctrack, which efficiently addresses the identity-switch problem in multi-object tracking using unfalsified control.

\begin{figure*}[!t]
\centering
\includegraphics[width=6in]{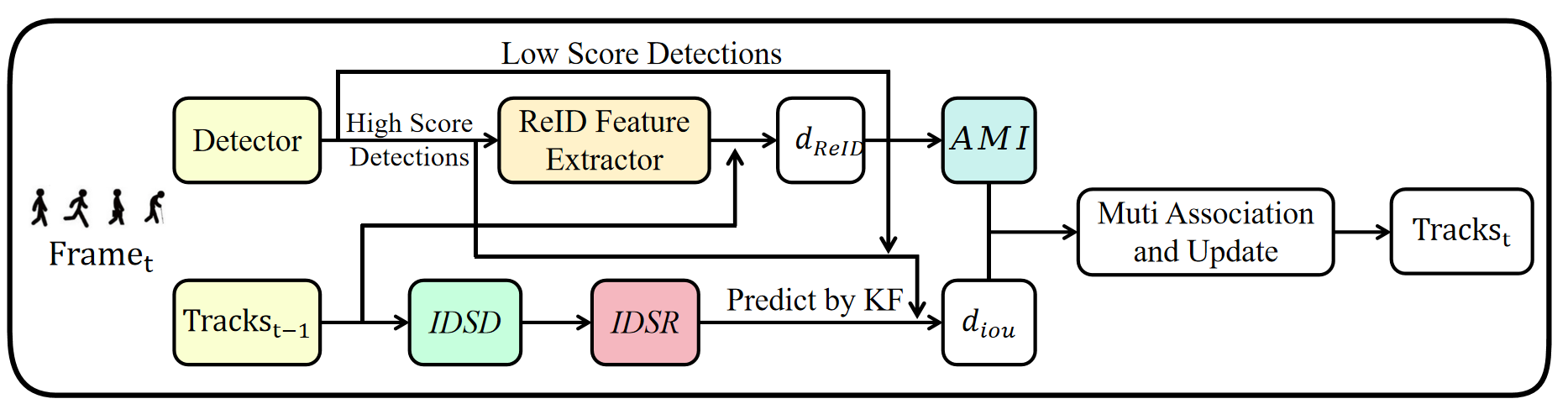}
\caption{Overview of the Unfctrack tracker's simplified channel diagram.}
\label{fig_framework}
\end{figure*}

\begin{figure*}[!t]
\centering
\includegraphics[width=6in]{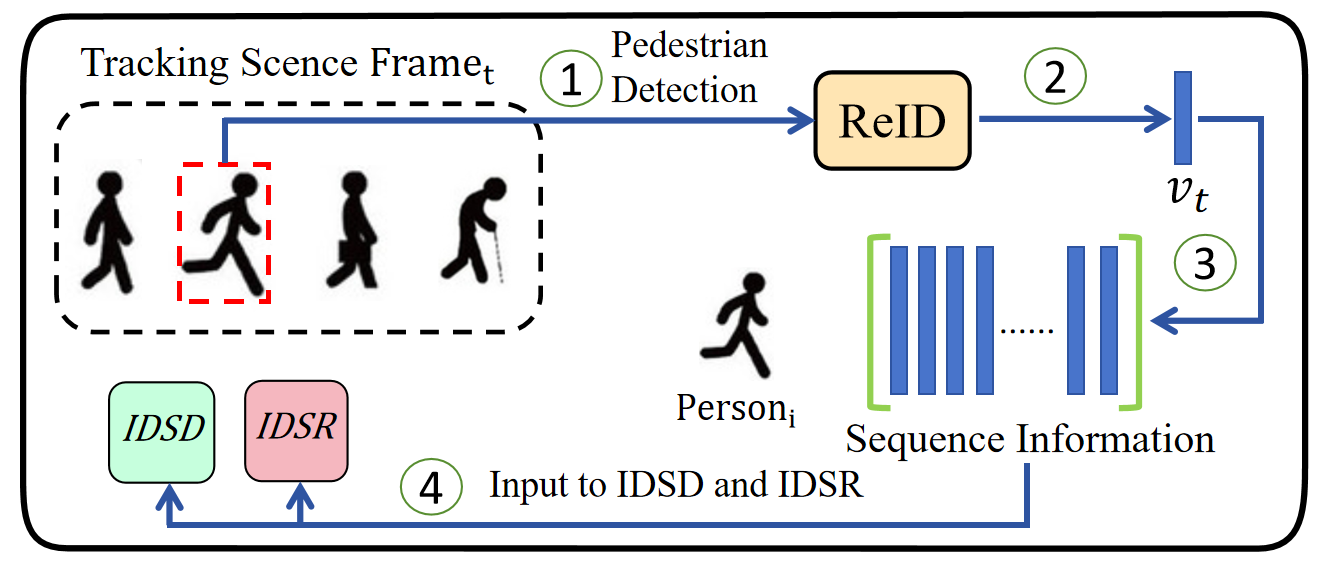}
\caption{The key to identifying identity switch lies in establishing a time series of appearance information, which is used for subsequent IDSD and IDSR modules.}
\label{fig_1}
\end{figure*}

\section{Methods}
In this section, we discuss the detailed design of our approach. Figure 1 represents the
architecture and workflow of our approach. Specifically, we aim to clearly illustrate the following three modules: 1)  ID-switch detection (IDSD) module, 2) ID-switch rectification (IDSR) module, and 3) ambiguous match improvement (AMI) module.

\vspace{-5pt}
\subsection{ID-switch detection (IDSD)} 
\vspace{-5pt}
Consider the state of the tracked object trajectory $O_i$ at time t=i-1 as $x_{i-1}$, and the state of the object within the detection box $K_i$ measured at time t=i as ${\hat{x}}_i$. We define the cost error as:
\begin{equation}
\label{deqn_ex1}
e_{K_i}^{O_i}=||{\hat{x}}_i-x_{i-1}||
\end{equation}
In the data association process between the current frame's tracked object trajectories and detection boxes, data association is only considered when $e_{K_i}^{O_i}$ is less than a given threshold $\epsilon$. Due to the existence of errors, multiple detection boxes may match with the same object trajectory. We include these multiple detection boxes in the candidate set $K$ for trajectory matching.
Suppose at time t=i, due to occlusion, there are two detection boxes in the candidate set K that match with the trajectory, $K$=[$k_1$, $k_2$], and the correct tracked target in the current frame is $k_2$. However, we select $k_1$ with a smaller error for data association, leading to an ID-switch. When the occlusion ends, the appearance information extracted based on the detection box becomes more accurate. As the ID-switch occurs, the difference between the historical appearance information stored in the trajectory and the appearance information in the current frame will gradually increase as is shown in Figure 2, and with time, it will exceed a threshold. Thus, we can determine that the trajectory has experienced an ID-switch. Therefore, the establishment of our unfalsified control model is as follows:
\begin{itemize}
\item Measurement information $P_{data}$: Appearance information extracted from objects during the tracking process.

\item Candidate set $K$: Detection boxes used for data association with trajectories.

\item Performance metric $T_{spec}$: The degree of change in appearance information of trajectories in a period of time.
\end{itemize}

To obtain measurement information $P_{data}$, we employ the BoT (SBS) \cite{ref29} feature extraction method from the FastReID \cite{ref20} library. We save appearance features every 5 frames, resulting in a sequence of appearance information with a length of 30. This sequence allows us to observe changes in appearance information. Even in the presence of occlusion-induced variations, the appearance information remains highly similar to the pre-occlusion state, providing data support for ID-switch detection. For the saved queue of appearance features, we establish another queue for post-processing to store the similarity costs between the current appearance feature and the previous appearance feature sequence. We calculate the cosine distance to measure the similarity between different appearance information. Then, we compute the variance of the similarity costs, which serves as the performance metric $T_{spec}$ for judging ID-switch cases.

We select the appearance feature $f$ of the current frame and calculate the cosine cost $C$ between $f$ and the previous 2⁄3 appearance features $f_i$ in the trajectory queue. We establish a cosine cost queue with a length of 30, and the calculation formula for $C$ is as follows:
\vspace{-5pt}
\begin{equation}
\label{deqn_ex1}
C\ =(\sum_{i=1}^{n}{(1-\frac{f*f_i}{||f||*||f_i||}}))/n
\end{equation}

Based on the values of C in the queue, we calculate $T_{spec}$, where $\overline{C}$ represents the average of $C_i$:
\vspace{-5pt}
\begin{equation}
\label{deqn_ex1}
T_{spec}=\frac{\sum_{i}^{n}{(C_i-\overline{C})}^2}{n}
\end{equation}
When there are short-term variations in appearance features due to occlusion, the variance will remain at a low level. However, when an ID-switch occurs and the appearance information undergoes long-term changes, the performance metric $T_{spec}$ will continue to rise. Once it exceeds a threshold $T_\theta$, reaching the falsification criterion, we consider the trajectory to have undergone an ID-switch and remove the matched trajectory from the candidate set $K$.
\vspace{-8pt}
\subsection{ID-switch rectification (IDSR)}
\vspace{-5pt}
After falsifying the trajectories that have undergone an ID-switch using unfalsified control, we attempt to restore the true detection boxes corresponding to these trajectories. We establish a queue for each trajectory to store its appearance information. Therefore, we can select the appearance information extracted before the occurrence of the ID-switch, which represents the appearance information when the trajectory was not subjected to an ID-switch, as data support for ID correction.

To ensure the accuracy of the rectification process, we consider a trajectory to be associated with a detection box and update the trajectory only when the cosine cost is below a very small threshold, denoted as $C_\theta$. In other words, the two appearance feature vectors are nearly identical. Furthermore, if we cannot find a suitable detection box to match the trajectory undergoing an ID-switch, we assign a new ID to the trajectory. This prevents it from continuing to be tracked using the incorrect ID.
\begin{figure*}[!t]
\centering
\includegraphics[width=6in]{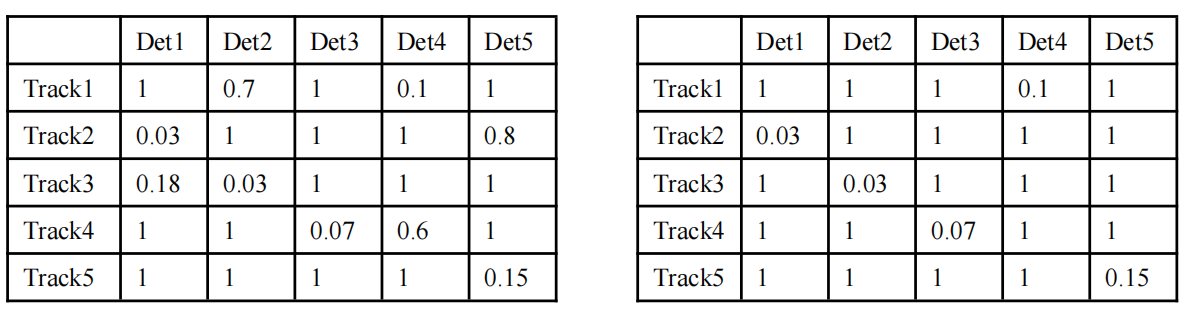}
\caption{AMI module addressing the ambiguous match issue between trajectories and detection boxes. Firstly, the module discards low-confidence matches with a threshold greater than $d_\theta$=0.2, setting them to 1. For each row and column, the module also discards low-weight ambiguous matches. For example, Track3-Det1 and Track2-Det1 are compared, and since 0.18 is significantly greater than 0.03, the module discards the match with a weight of 0.18 and sets it to 1.}
\label{fig_2}
\end{figure*}

\vspace{-15pt}
\subsection{Ambiguous match improvement (AMI)}
\vspace{-5pt}
The fusion of motion information and appearance information is performed in a similar manner to \cite{ref7}, using a weighted fusion approach. The motion information, represented by $d_{iou}$, and the appearance information, represented by $d_{ReID}$, are fused to obtain the fused information $d_{dist}$, using the following fusion method:
\begin{equation}
\label{deqn_ex1}
d_{dist}=\alpha*d_{iou}+(1-\alpha)*d_{ReID}(0<\alpha<1)
\end{equation}
The appearance information $d_{ReID}$, is obtained by extracting features from the detection boxes. However, many detection boxes suffer from overlap issues, resulting in significant ambiguity in some $d_{ReID}$ values. This ambiguity can lead to severe fuzzy matching problems, potentially causing ID-switch and even discarding tracked trajectories for further processing in the next frame. To address this issue, we have designed a simple yet effective module called AMI to handle the problem of fuzzy matching. Specifically, we discard matches with confidence scores higher than a threshold $d_\theta$, for low-confidence matches. For each row and column, we calculate the weights for high-confidence matches and low-confidence matches, respectively, and discard the low-confidence matches with lower weights.

\begin{figure*}[!t]
\centering
\includegraphics[width=6in]{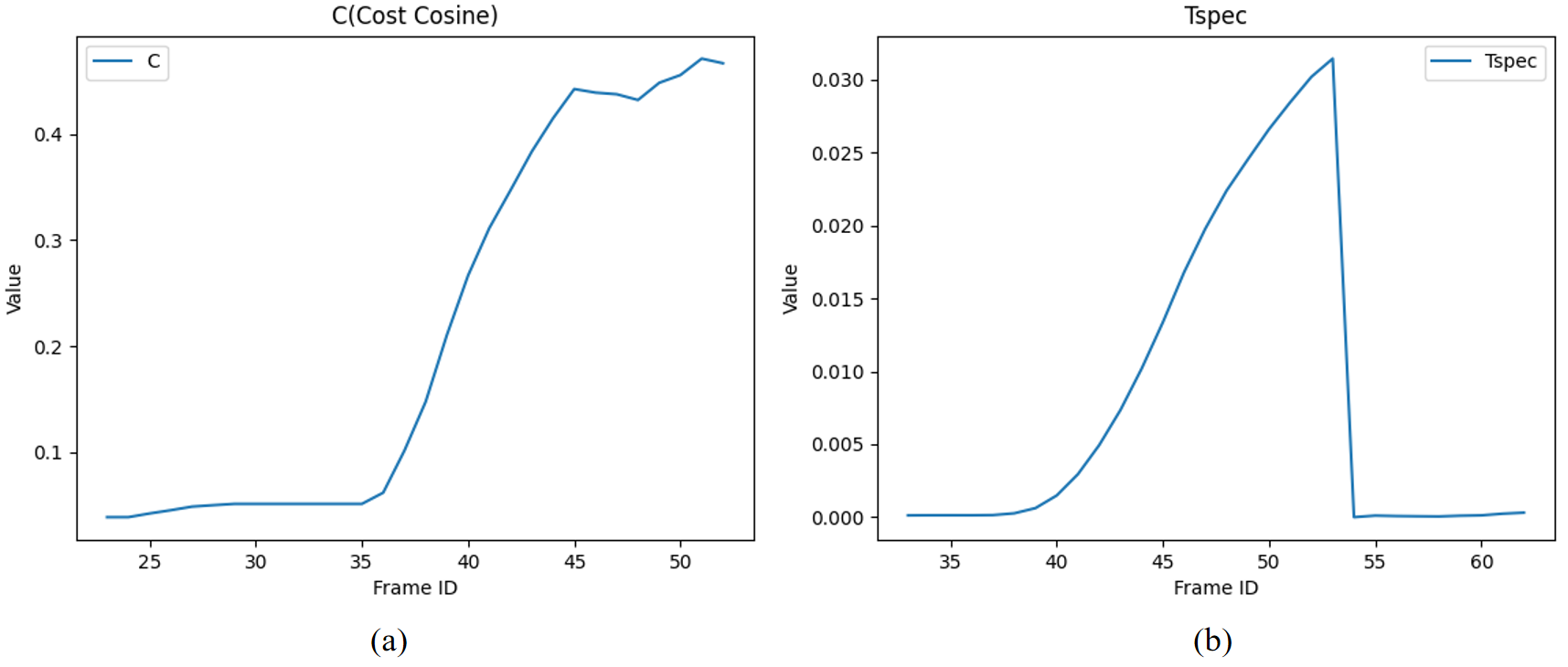}
\caption{Experimental results of IDSD and IDSR on the MOT17-01 dataset. The two subfigures illustrate the increasing cosine cost and $T_{spec}$ after an ID-switch occurrence, as well as the variation of $T_{spec}$ after ID rectification. (a)The change of cosine cost.(b)The change of $T_{spec}$.}
\label{fig_5}
\end{figure*}

\begin{figure*}[!t]
\centering
\includegraphics[width=6in]{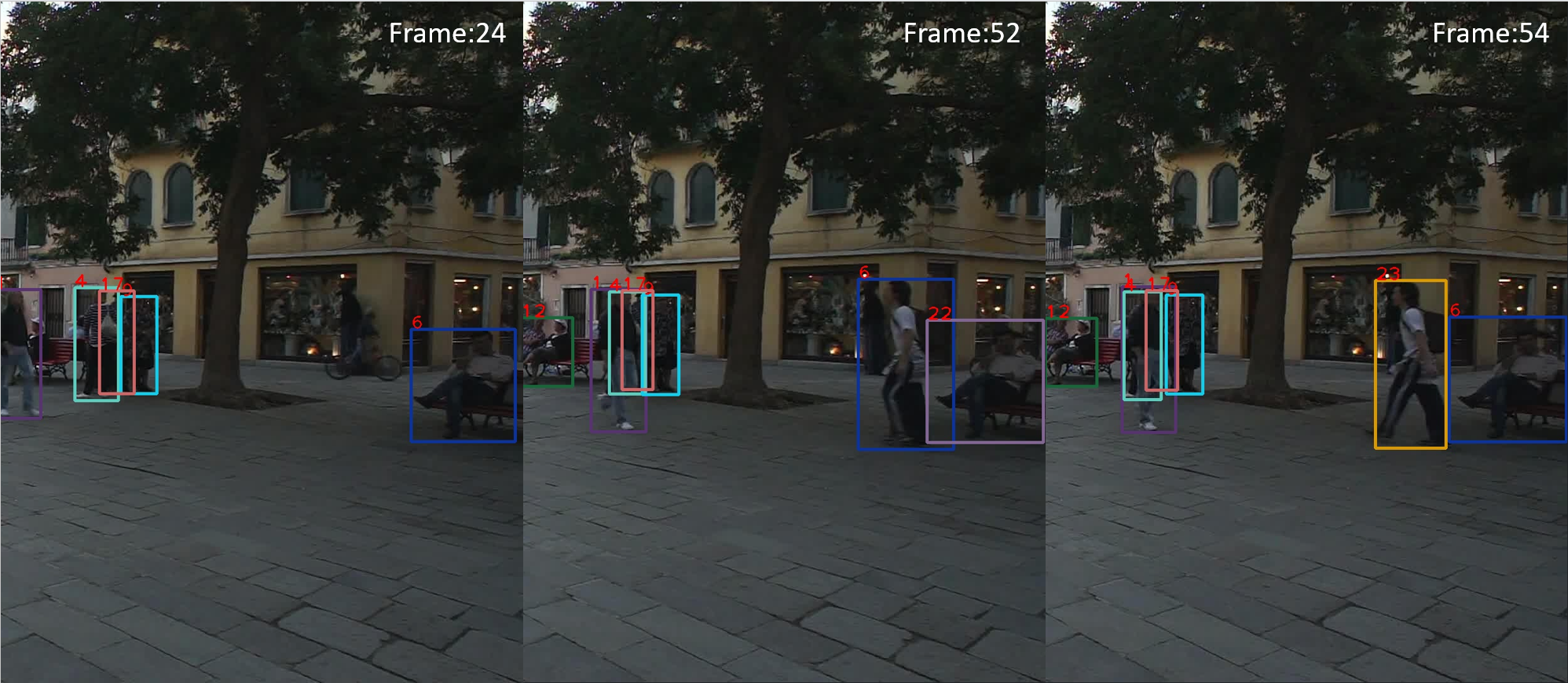}
\caption{The three subfigures show the experimental image results at frames 24, 52, and 54.The IDSD model determines the ID-switch at frame 53, and the IDSD model rectify the ID at frame 54.}
\label{fig_5}
\end{figure*}

\vspace{-20pt}
\section{Experiments}
\vspace{-5pt}
\subsection{Experimental settings}
\vspace{-5pt}
\noindent\textbf{Datasets.} The experiments are conducted on the MOT17 \cite{ref14} and MOT20 \cite{ref15} datasets under the "private detection" protocol. The MOT17 and MOT20 datasets are widely used benchmarks in the field of multi-object tracking. Since both MOT17 and MOT20 do not provide a separate validation set, we follow the common practice \cite{ref41,ref3} of splitting the training set into halves for training and validation.

\noindent\textbf{Metrics.} Existing methods in the field have not considered the possibility of recovering IDs after an ID-switch occurs. Therefore, widely accepted metrics in multi-object tracking, such as MOTA, HOTA, and IDSW \cite{ref36}, are not suitable for evaluating the performance of our tracker. We observe in our experiments that if the ID of a detected object changes once, IDSW increases by 1. However, if we correct the IDs of objects that have undergone an ID-switch, IDSW increases by 1 instead of decreasing by 1. This results in an increase in IDSW and a decrease in MOTA when we judge and correct ID-switch situations. Therefore, metrics like MOTA and IDSW cannot effectively evaluate our tracker, and we emphasize the potential of addressing the ID-switch problem from different perspectives.We demonstrate the effectiveness of our IDSD and IDSR modules through four metrics: RC (Recall, the proportion of IDSW correctly detected by the IDSD module), FPR (False Positive Rate, the proportion of IDSW incorrectly detected by the IDSD module), RA (Rectification Accuracy, the proportion of IDSW correctly rectified by the IDSR module), and RR (Renumber Ratio, the proportion of IDSW renumbered by the IDSR module).

\vspace{-2pt}
\noindent\textbf{Implementation details.} Our multi-object tracking method follows the detection-to-tracking paradigm. We use the publicly available YOLOX \cite{ref37} as the detector, and the detection results are then used as input for our tracker. For ReID feature extraction, we utilize the SBS-50 model from the open-source FastReID \cite{ref20}, which is pre-trained on MOT17 and MOT20 \cite{ref2}. Throughout the experiments, we set the default detection score threshold $\tau$ to 0.6 and remove a trajectory if it is lost for more than 30 frames. The threshold value for $T_\theta$ in IDSD is set to 0.01 by default, and the threshold value for the cosine cost $C_\theta$ in IDSR is set to 0.1 by default.

\vspace{-15pt}
\subsection{Testing and experiments results}
\vspace{-5pt}
In order to demonstrate the tracker's ability in detecting and correcting ID-switches, the AMI module is not used during the experimentation with the IDSD and IDSR modules.

\noindent\textbf{ID-switch detection(IDSD).} In the MOT17-01 dataset, we set the threshold $T_\theta$ for $T_{spec}$ in IDSD to 0.01 to evaluate the tracker's ability to detect ID-switch situations. To exclude temporary variations in trajectory appearance features due to occlusion rather than ID-switch, we consider a trajectory to have undergone an ID-switch only when its performance metric $T_{spec}$ exceeds $T_\theta$ for more than 10 consecutive frames. In the experiment, at frame 40, a new object enters from the right side of the field of view, resulting in an ID-switch for trajectory 6. We observe that the performance metric $T_{spec}$ for trajectory 6 exceeds $T_\theta$ at frame 43, and the tracker determines the occurrence of an ID-switch for trajectory 6 at frame 53 which is shown in Figure 5.

\noindent\textbf{ID-switch rectification (IDSR).} In the MOT17-01 dataset, after falsifying trajectory 6 for an ID-switch at frame 53 using unfalsified control, we retrieve the first appearance feature $f_1$ from the trajectory 6 queue. This feature $f_1$ was extracted before the ID-switch occurred. We match this feature with the appearance information of the current frame's detection box, calculate the cosine cost, and if the cosine cost is below the threshold $C_\theta$, we perform ID rectification. Simultaneously, we remove the erroneous trajectory 23 and assign a new ID, 24, to the newly entered object on the right side. We observe that after correctly recovering the ID, the performance metric $T_{spec}$ decreases below the threshold $T_\theta$ and stabilizes.

In the experiments conducted on other datasets in MOT17 and MOT20, our tracker demonstrates excellent ability in falsifying trajectories that have undergone ID-switch. It is worth noting that due to differences in camera angles and tracking environments across different datasets, there may be variations in setting the threshold $T_\theta$ for $T_{spec}$ in IDSD.In the MOT17 test set, we validated the effectiveness of the IDSD and IDSR modules and presented the results in Table 1. In the experiment, the IDSD module was able to detect nearly half of the IDSW conditions (51.62 percent) with a relatively low false positive rate. The IDSR module was able to correct 23.1 percent of the IDSW and renumber the uncorrectable targets to prevent affecting the tracking results.

\noindent\textbf{Ambiguous match improvement (AMI).} We conduct ablation experiments on the AMI module using the training sets of MOT17 and MOT20. We evaluate the tracker using the official evaluation tool, Trackeval, from the MOT challenge. We find that applying the AMI algorithm to process the appearance information $d_{ReID}$, resulted in significant performance improvement in tracking.
\begin{table*}[htbp]
  \centering
  \begin{tabularx}{\linewidth}{>{\raggedright\arraybackslash}p{2.5cm}|XXXXX}
    \toprule
    MOT17-test & IDSW & RC & FPR & RA & RR\\
    \midrule
    IDSD & 554    & 51.62  & 5.3    & \multicolumn{1}{p{0.5cm}}{/} & \multicolumn{1}{p{0.5cm}}{/}\\
    IDSR & 554  & \multicolumn{1}{p{0.5cm}}{/}  & \multicolumn{1}{p{0.5cm}}{/}  & 23.1 & 76.9 \\
    \bottomrule
  \end{tabularx}
  \captionsetup{font=small,labelsep=colon} 
  \caption{Metrics of IDSD and IDSR modules in the MOT17 test set.}
  \label{tab:addlabel}
\end{table*}

The experiments primarily emphasize the results of ID-switch detection and rectification. Although the metrics from \cite{ref36} are not suitable for effectively evaluating our tracker, we still provide experimental results for the unfctrack tracker(without IDSD and IDSR modules) on the MOT17 dataset in Table 3. 

\vspace{-10pt}
\subsection{Limitations}
\vspace{-5pt}
Unfctrack heavily relies on appearance feature information for the detection and recovery of ID-switch situations. However, excessive ReID feature extraction in dense scenes can be time-consuming, potentially compromising real-time performance. Additionally, the effectiveness of ReID feature extraction significantly affects the performance of IDSD and IDSR. Factors such as camera movement and significant environmental background variations may cause certain parameters of the unfctrack tracker to change. For example, the threshold value $T_\theta$ for the performance metric $T_{spec}$ in IDSD may differ across different environments. For instance, $T_\theta$ = 0.01 may be suitable for MOT17-01, while $T_\theta$ = 0.02 may be more appropriate for MOT17-03. Adjustments may be necessary for $T_\theta$ in different environments. Finally, although Unfctrack demonstrates strong capabilities in detecting ID-switch situations, the correction and recovery of IDs still present significant challenges, which will be a focus of future research efforts.

\begin{table}[htbp]
  \centering
  \resizebox{\columnwidth}{!}
  {
    \begin{tabular}{c|rrrr}
        \toprule
        Tracker & \multicolumn{1}{c}{MOTA↑} & \multicolumn{1}{c}{FP↓} & \multicolumn{1}{c}{FN↓} & \multicolumn{1}{c}{IDSW↓} \\
        \midrule
        MOT17 & 87.492 & 3355  & 10162 & 529 \\
        MOT17(AMI) & 90.042 & 2282  & 8540  & 361 \\
        MOT20 & 89.456 & 12703 & 37689 & 1093 \\
        MOT20(AMI) & 92.762 & 6315  & 28430 & 594 \\
        \bottomrule
    \end{tabular}%
  }
  \captionsetup{font=small,labelsep=colon} 
  \caption{Validation of the AMI module results on the train datasets of MOT17 and MOT20 using the official evaluation tool Trackeval from MOT-challenge.} 
  \label{tab:addlabel}%
\end{table}

\begin{table*}[htbp]
  \centering
  \begin{tabularx}{\linewidth}{>{\raggedright\arraybackslash}p{2.5cm}|XXXXXXX}
    \toprule
    Tracker & MOTA↑ & IDF1↑ & HOTA↑ & FP↓ & FN↓ & IDs↓ & FPS↑ \\
    \midrule
    Tube-TK \cite{ref30} & 63    & 58.6  & 48    & 27060 & 177483 & 4137  & 3 \\
    GSDT \cite{ref1} & 66.2  & 68.7  & 55.5  & 43368 & 144261 & 3318  & 4.9 \\
    LMOT \cite{ref32} & 72    & 70.3  & 56.7  & 28113 & 126704 & 3071  & 28.6 \\
    MOTR \cite{ref8} & 73.4  & 68.6  & 57.8  & \multicolumn{1}{p{0.5cm}}{/} & \multicolumn{1}{p{0.5cm}}{/} & 2439  & \multicolumn{1}{p{0.5cm}}{/} \\
    FairMOT \cite{ref11} & 73.7  & 72.3  & 59.3  & 27507 & 117477 & 3303  & 25.9 \\
    Transtrack \cite{ref19} & 75.2  & 63.5  & 54.1  & 50157 & 86442 & 3603  & 59.2 \\
    CrowdTrack\cite{ref33} & 75.6  & 73.6  & 60.3  & 25950 & 109101 & 2544  & 140.8 \\
    STC \cite{ref34} & 75.8  & 70.9  & 59.8  & 44952 & 87039 & 4533  & 9.5 \\
    FCG \cite{ref35} & 76.7  & 77.7  & 62.6  & 13284 & 116205 & 1737  & 4.9 \\
    OC-SORT \cite{ref5} & 78    & 77.5  & 63.2  & 15129 & 107055 & 1950  & 29 \\
    Bytetrack \cite{ref3} & 80.3  & 77.3  & 63.1  & 25491 & 83721 & 2196  & 29.6 \\
    Unfctrack(ours) & \textbf{79.8}  & \textbf{77.9}    & \textbf{63.5}    & 21960 & 90834 & \textbf{1662}  & 8.3 \\
    \bottomrule
  \end{tabularx}
  \captionsetup{font=small,labelsep=colon} 
  \caption{Performance metrics comparison of Unfctrack tracker and other trackers on the MOT17 dataset under the private detection protocol.Due to the metrics not being suitable for measuring our IDSD and IDSR processes, Unfctrack did not use these two modules in the results.}
  \label{tab:addlabel}
\end{table*}
\vspace{-10pt}
\section{CONCLUSIONS}
\vspace{-5pt}
In this paper, we propose unfctrack, which utilizes unfalsified control to identify and attempt to rectify ID-switch situations. By leveraging data-driven unfalsified control, our tracker can dynamically identify and rectify errors during the tracking process. To the best of our knowledge, this is the first tracker that focuses on detecting and attempting to correct ID-switch occurrences. The proposed method, which incorporates appearance information, can be easily integrated into other tracking frameworks. We hope that this work provides a new perspective for addressing ID-switch problems and contributes to the advancement of the field of multi-object tracking.


\bibliographystyle{unsrt}
\bibliography{example}

\end{document}